\documentclass{article}

\usepackage[final]{corl_2018} %

\newcommand{\lidar}{LiDAR}

\title{Learning to Localize Using a \lidar{} Intensity Map}

\usepackage{booktabs}	%
\usepackage[inline]{enumitem}   %
\usepackage{graphics} %
\usepackage{epsfig}   %
\usepackage[utf8]{inputenc} 	%
\usepackage{amsmath}  %
\usepackage{amssymb}  %
\usepackage[colorinlistoftodos,prependcaption]{todonotes} %
\usepackage{subcaption}
\usepackage{adjustbox}
\usepackage{algorithm}
\usepackage[noend]{algpseudocode}
\usepackage{microtype}    %

\newcommand{\fref}[1]{Fig.~\ref{#1}}
\newcommand{\tref}[1]{Table~\ref{#1}}

\newcommand{\slwang}[1]{\textcolor{black}{#1}}
\newcommand{\andreibc}[1]{\textcolor{black}{#1}}

\newcommand{\bx}{\mathbf{x}}

\newcommand{\bz}{\mathbf{z}}
\newcommand{\bp}{\mathbf{p}}

\newcommand{\bw}{\mathbf{w}}
\newcommand{\cI}{\mathcal{I}}
\newcommand{\cF}{\mathcal{F}}
\newcommand{\cL}{\mathcal{L}}
\newcommand{\cG}{\mathcal{G}}
\newcommand{\cM}{\mathcal{M}}

\newcommand{\ut}{^{(t)}}

\newcommand{\mlidar}{\text{\lidar}}
\newcommand{\mgps}{\text{GPS}}

\newcommand{\online}{\textsc{o}}
\newcommand{\map}{\textsc{m}}

\newcommand{\degree}{$^{\circ}$}

\author{
  Ioan Andrei B\^{a}rsan$^{\ast,1, 2}$\ \    Shenlong Wang$^{\ast,1, 2}$\
  \     Andrei Pokrovsky$^{1}$\ \    Raquel Urtasun$^{1, 2}$ \\
$^1$Uber ATG, \  $^2$University of Toronto\\
\texttt{\{andreib, slwang, andrei, urtasun\}@uber.com} \\
}

\begin{document}
\maketitle

\begin{abstract}
In this paper we propose a real-time, calibration-agnostic and effective localization system for self-driving cars.  Our method learns to embed the online \lidar{} sweeps and intensity map into a joint deep embedding space. 
Localization is then conducted through an efficient convolutional matching
between the embeddings.  Our full system can operate in real-time at 15Hz  while
achieving centimeter level accuracy across different \lidar{} sensors and
environments.  Our  experiments illustrate the performance of the proposed
approach over a large-scale dataset consisting of over 4000km of driving.
\end{abstract}


\section{Introduction}
	
One of the fundamental problems in autonomous driving is to be able to
accurately localize the vehicle in real time.
Different precision requirements exist depending on
\andreibc{the intended use of the localization system.}
For routing the self-driving vehicle  from point A to
point B, precision of a few meters is sufficient. However, centimeter-level
localization  becomes necessary in order to exploit high definition (HD) maps
as priors for robust perception, prediction\andreibc{,} and safe motion planning.

Despite many decades of research, reliable and accurate localization remains an
open problem, especially when very low latency is required.
Geometric methods, such as those based on the iterative closest-point algorithm
(ICP)~\cite{besl1992method, rusinkiewicz2001efficient} can lead to high-precision localization, but remain
vulnerable in the presence of  geometrically non-distinctive or repetitive environments, such as
tunnels, highways, or bridges. %
Image-based methods~\cite{cummins2008fab, ziegler2014video,
arandjelovic2016netvlad, Sattler2017} are
also  capable of robust localization, but are still behind geometric ones in
terms of outdoor localization precision. Furthermore, they often require
capturing the environment in different seasons and times of the day as the appearance might change dramatically. 

A promising alternative to these  methods is to leverage LiDAR intensity
maps~\cite{levinson2007map, levinson2010robust}, which encode information
about the appearance and semantics  of the scene.  However, the intensity of
commercial LiDARs is inconsistent across different beams and manufacturers, and
prone to changes due to environmental factors such as temperature. 
Therefore, intensity based localization methods rely heavily on having very accurate intensity calibration of each LiDAR beam.  This requires careful
fine-tuning of each vehicle to achieve good performance, sometimes
\andreibc{even on} a daily basis. Calibration can be a very laborious process,
\andreibc{limiting the scalability of this approach}.
Online calibration is a promising solution, but current approaches  fail to
deliver the desirable accuracy.  
Furthermore, maps have to be re-captured each time we change the sensor, e.g.,
\andreibc{to exploit a new generation of \lidar{}}.

In this paper, we  address the aforementioned problems  by learning to perform intensity based localization. 
Towards this goal, we design a deep  network that embeds both   LiDAR intensity maps and  online LiDAR sweeps in a common space where calibration is not required. 
Localization is then simply done by searching exhaustively over 
\andreibc{3-DoF
poses (2D position on} the map manifold plus rotation), where the score of each
pose can be computed by the cross-correlation between the embeddings. This allows us
to perform localization in a few 
\andreibc{milliseconds}
on the GPU. %

We demonstrate the effectiveness of our approach in both highway and urban
environments over 4000km of roads. %
Our experiments showcase  the  advantages of our approach over traditional methods, such as the
ability to work with uncalibrated data \andreibc{and} the ability to generalize across
different \lidar{} sensors.%

\section{Related Work}

\paragraph{Simultaneous Localization and Mapping:}
Given a sequence of sensory inputs (e.g., LiDAR point clouds, color and/or depth images)  simultaneous localization and mapping (SLAM) approaches \cite{lsdslam, orbslam, jizhang} reconstruct a map of the environment and estimate the relative  poses between each  input and the map.  Unfortunately, since the estimation error is usually biased, accumulated errors 
cause gradual estimation drift
as the robot moves, resulting in large errors. 
Loop closure has been largely used to fix this issue. However, in many
scenarios such as highways, it is unlikely that trajectories are closed.
GPS measurements \slwang{can help reduce the drift issue through fusion but
commercial GPS sensors \andreibc{are not able to achieve centimeter-level}
accuracy.} %

\paragraph{Localization Using Light-weight Maps:} %
\slwang{Light-weight maps, such as Google maps and OpenStreetMap, draw attention for developing affordable localization efforts. 
While only requiring small amounts of storage, they encode both the topological
structure of the road network, as well as its semantics. Recent approaches incorporated
them to compensate for large drift \cite{brubaker2013lost, ma2016find} and keep the
vehicle associated with the road.}
\andreibc{However, these methods are still not yet able to achieve
  centimeter-level accuracy.}

\paragraph{Localization Using High-definition Maps:}
Exploiting high-definition maps (HD maps)  has gained attention in recent years
on both indoor and outdoor localization  \cite{ziegler2014video, 
  levinson2007map, levinson2010robust, rwolcott2015gmm, wolcott2014visual, 
  schreiber2013laneloc, yoneda2014lidar, Wan2017}. The general idea is to build an accurate
map of the environment offline  through aligning multiple sensor passes over
the same area. In the online stage the system is able to  achieve sub-meter level accuracy by matching the sensory input against the
HD-map. 
\andreibc{In their 
pioneering work, \citet{levinson2007map} built a \lidar{} intensity map offline 
using Graph-SLAM \cite{thrun2006graph} and used particle filtering and Pearson
product-moment correlation to localize against it. In a similar
fashion,~\citet{Wan2017} use BEV \lidar{} intensity images in conjunction with
differential GPS and an IMU to robustly localize against a pre-built map, using
a Kalman filter to track the uncertainty of the fused measurements over time.}
Uncertainty in intensity changes can be handled through online calibration and
by building probabilistic  map priors \cite{levinson2010robust}. However, these
methods  require accurate calibration (online or offline) which is difficult to
achieve. \citet{yoneda2014lidar} proposed to align online \lidar{} sweeps against 
an existing 3D prior map using ICP. However, this approach suffers  in the presence of  repetitive geometric structures, such as highways and bridges. 
The work of \citet{rwolcott2015gmm} combines height and intensity information against a GMM-represented height-encoded map and accelerates registration using branch and bound. Unfortunately, the inference speed cannot satisfy the real-time requirements of self-driving. 
Visual cues from a camera can also  be  utilized to match against 3D prior 
maps~\cite{cummins2008fab, ziegler2014video,wolcott2014visual}. However, these approaches either require computationally demanding online 3D map rendering \cite{wolcott2014visual} or 
\slwang{\andreibc{lack robustness} to visual apperance changes due to the time
of day, weather, and seasons}~\cite{cummins2008fab, ziegler2014video}.
Semantic cues such as lane markings \andreibc{can also be used to build a} 
prior map and exploited for localization \cite{schreiber2013laneloc}. 
However, \andreibc{the effectiveness of such methods} depends on perception
performance and does not work for regions 
\andreibc{where such cues are absent}.

\begin{figure}[t]
  \centering
  \includegraphics[width=0.95\textwidth]{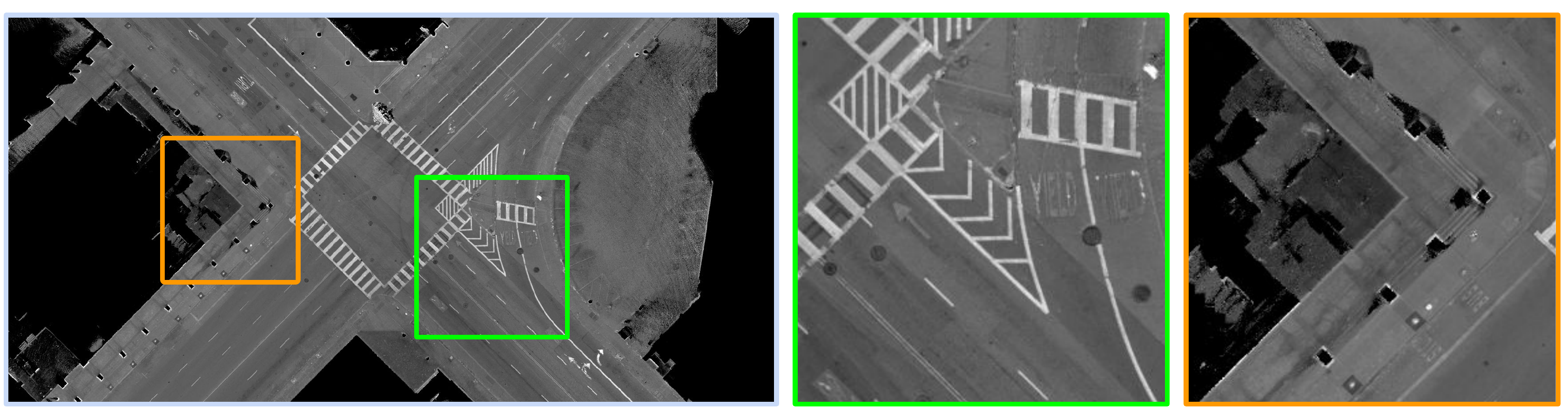}  
  \caption{An example of a bird's eye view (BEV) \lidar{} intensity map used by
  our system. It encodes rich information on both appearance and geometry
structure information for localization. The orange square highlights an example
of geometric structure captured by the BEV images, while the green one
highlights an example of intensity structure.\label{fig:map-example}}
\end{figure}

\paragraph{Matching Networks:} Convolutional matching networks that compute
similarity between local patches have been exploited for tasks such as stereo   \cite{zbontar2015computing, luo2016efficient,han2015matchnet}, flow estimation  \cite{xu2017accurate, bai2016exploiting}, global feature correspondence
\cite{zagoruyko2015learning}, and 3D \slwang{voxel matching} \cite{zeng20173dmatch}. 
In this paper, we extend this line of work for \andreibc{the task of localizing
in a known map.}

\paragraph{Learning to Localize:}
Training  machine learning models to conduct self-localization is an emerging field in robotics. 
The pioneer\andreibc{ing} work of \citet{shotton2013scene} trained random forest 
to detect corresponding local features between a single depth image and a 
pre-scanned indoor 3d map. \cite{WangICCV15} utilized CNNs to detect text from large shopping malls to conduct indoor localization.  
\citet{kendall2015posenet} proposed to directly regress a 6-DoF pose from a 
single image. 
Neural networks have also been  used to learn representations for place recognition in outdoor scenes, achieve state-of-the-art performance in localization-by-retrieval \cite{arandjelovic2016netvlad}. Several algorithms utilize deep learning for end-to-end visual odometry \cite{zhou2017unsupervised, wang2017deepvo}, 
\andreibc{showing promising results but still remaining behind traditional SLAM
approaches in terms of performance}.

Very recently, \citet{bloesch2018codeslam} learn a CNN-based feature
representation from intensity images to encode depth, which is used to conduct
incremental structure-from-motion inference. \andreibc{At the same time, the methods of
\citet{Brachmann2017} and \citet{Radwan2018} push the state of the art in
learning-based localization to impressive new levels, reaching centimeter-level accuracy
in indoor scenarios, such as those in the 7Scenes dataset, but not outdoor ones.}

\begin{figure}
  \centering
  \includegraphics[width=0.98\textwidth]{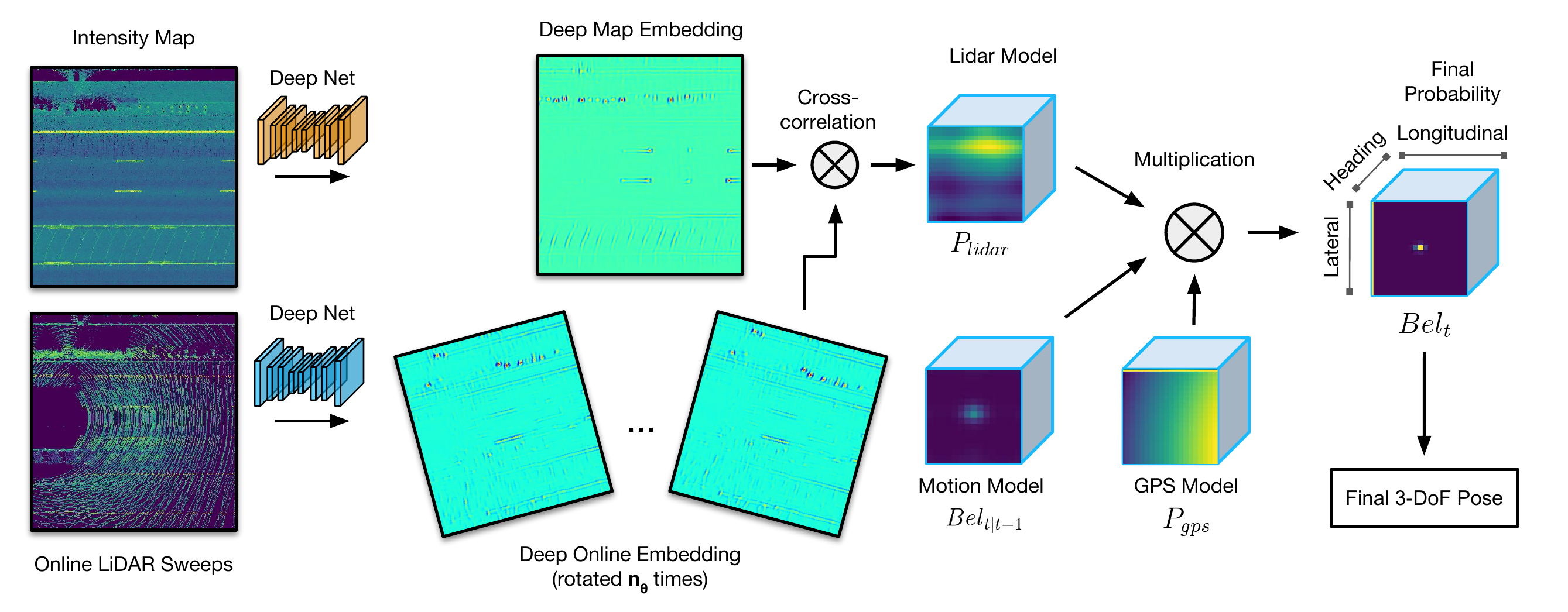}
  \caption{The full end-to-end architecture of our proposed deep localization network.}\label{fig:teasr}
\end{figure}

\section{Robust Localization Using \lidar{} Data}
\label{sec:robust-localization}

\newcommand{\cudnn}{cuDNN}

In this section, we discuss our %
\lidar{}  intensity 
localization system.  %
We first formulate
localization as \slwang{deep recursive Bayesian estimation} problem and discuss
each probabilistic term. We then present our real-time inference algorithm
followed by a description of how our model is trained. %

\subsection{\lidar{} Localization as a Deep \andreibc{Recursive} \slwang{Bayesian} Inference}

We perform high-precision localization against pre-built \lidar{} intensity
maps. The maps are constructed from multiple passes through the same area,
which allows us to perform additional post-processing steps, such as dynamic
object removal. The accumulation of multiple passes also produces maps which
are much denser than individual \lidar{} sweeps. The maps are encoded as
orthographic bird's eye view (BEV) images of the ground. We refer the reader
to~\fref{fig:map-example} for a sample fragment of the maps used by our
system.

Let $\bx$ be the pose of the
self-driving vehicle (SDV). We assume that our  sensors are 
calibrated and neglect the effects of suspension, unbalanced tires, and
vibration. This enables us to simplify the vehicle's 6-DoF pose to only 3-DoF, namely a 2D translation and a heading angle, i.e., $\bx =
\{x, y, \theta \}$, where $x, y \in \mathbb{R}$ and 
$\theta \in (-\pi, \pi]$. 
At each time step $t$, our \lidar{} localizer takes as input the previous
pose most likely estimate $\bx_{t-1}^\ast$ and uncertainty $\mathrm{Bel}_{t-1}(\mathbf{x})$, the vehicle dynamics $\dot{\bx}_t$, the online \lidar{}
image $\cI_t$, and the pre-built \lidar{} intensity map $\cM$. In
order to generate $\cI\ut$, we aggregate the $k$ most recent
\lidar{} sweeps using the IMU and wheel odometry. This
produces denser online \lidar{} images than just using the most recent sweep,
helping  localization. Since $k$ is small, drift is not an issue. 

We then  formulate the localization problem as {deep recursive bayesian inference}
problem. We encode the fact that the online \lidar{} sweep should be consistent
with the map at the vehicle's location\slwang{, consistent with GPS readings}
and \andreibc{that the belief updates should be consistent with the motion
model.} \slwang{Thus}

\begin{equation}
\mathrm{Bel}_t(\bx) = \eta \cdot P_{\mlidar}(\cI_t | \bx; \bw) P_{\mgps}(\cG_t | \bx) \mathrm{Bel}_{t|t-1} (\bx | \mathcal{X}_t) 
\end{equation}
where $\bw$ is a set of learnable parameters, $\cI_t$, $\cG_t$ and
$\mathcal{X}_t$ are the \lidar{}, GPS, and vehicle dynamics observation
respectively.
$\mathrm{Bel}(\mathbf{x}_t)$ is the posterior distribution of the vehicle pose
at time $t$ given all the sensor observations until step $t$; $\eta$ is
a normalization factor. We do not need to calculate it explicitly because we
discretize the belief space, so normalization is trivial.

\begin{figure}[t]
  \centering
  \begin{subfigure}[t]{0.24\textwidth}
    \centering
    \includegraphics[width=1.0\linewidth]{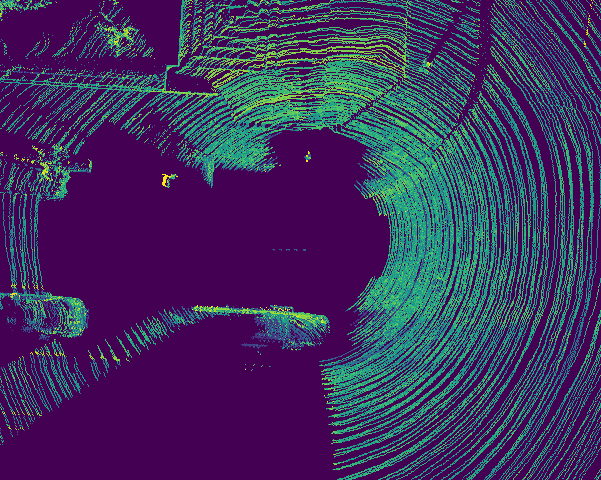}
    \caption{Online \lidar{} Image.}
  \end{subfigure}~
  \begin{subfigure}[t]{0.24\textwidth}
    \centering
    \includegraphics[width=1.0\linewidth]{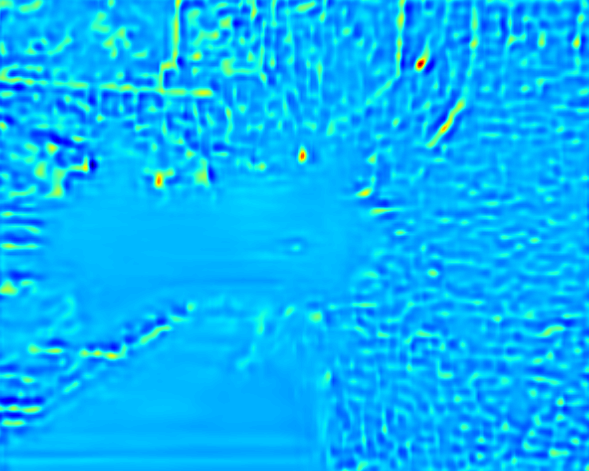}
  \caption{Online embedding.}
  \end{subfigure}~
  \begin{subfigure}[t]{0.24\textwidth}
    \centering
    \includegraphics[width=1.0\linewidth]{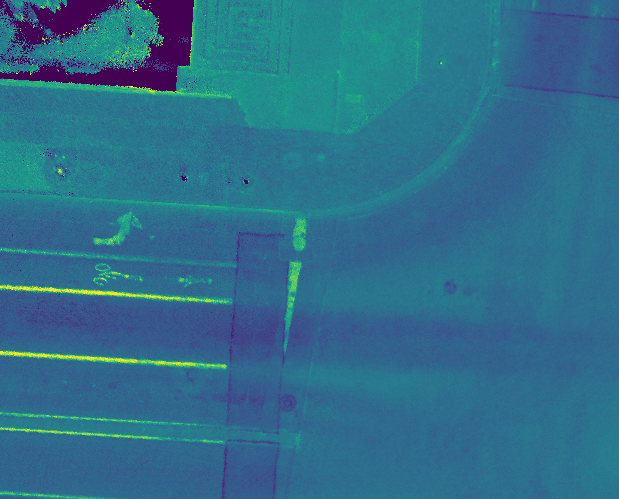}
    \caption{Intensity map.}
  \end{subfigure}~
  \begin{subfigure}[t]{0.24\textwidth}
    \centering
    \includegraphics[width=1.0\linewidth]{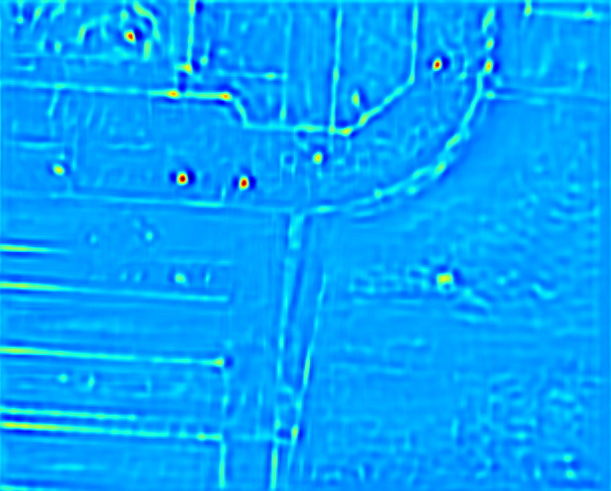}
  \caption{Map embedding.}
  \end{subfigure}~
  \caption{One example of the input and map embeddings.}
  \label{fig:embedding-example}
\end{figure}

\paragraph{\lidar{} Matching Model:} Given a candidate pose $\bx$, our
\lidar{} matching probability $P_{\mlidar}$ encodes the agreement between the current
online \lidar{} observation and the map indexed at the hypothesized 
pose $\bx$. To compute the probability, we first project both the map $\cM$ and the
online \lidar{} intensity image $\cI$ into an embedding space
using two deep neural networks. We then 
warp the online embedding according to each pose hypothesis, and
compute the cross-correlation between the warped online embedding and the
map embedding. Formally, this can be written as:
\begin{equation}
  P_{\mlidar} \propto s \left(\pi\left(f(\cI; \bw_\online), \bx\right), g(\cM; \bw_\map) \right),
\end{equation}
where $f(\cI; \bw_\online)$ and $g(\cM; \bw_\map)$ are the deep embedding networks of
the online \lidar{} image and the map, respectively, and
$\bw_o$ and $\bw_m$ are the networks' parameters.
$\pi$ represents a 2D rigid warping function meant to transform the online
\lidar{}'s embedding into the map's coordinate frame according to the given
pose hypothesis $\bx$. Finally, $s$ represents a cross-correlation operation.

Our embedding functions $f(\, \cdot \, ; \bw_\online)$ and 
$g(\, \cdot \,; \bw_\map)$ are customized fully convolutional  neural networks.
The first branch $f(\, \cdot\, ; \bw_\online)$ takes as input the bird's eye view
(BEV) rasterized image of the $k$ most recent \lidar{} sweeps (compensated by ego-motion) and produces a
dense representation at the same resolution as the input.
The second branch $g(\, \cdot \,; \bw_\map)$ takes as input a section of the 
\lidar{} intensity map, and produces an embedding with the same number of
channels as the first one, and the spatial resolution of the map.

\paragraph{GPS Observation Model:} The GPS observation model encodes the likelihood of GPS observation given a location proposal. We approximate uncertainty of GPS sensory
observation \andreibc{using a} Gaussian distribution:
\begin{equation}
P_{\mgps} \propto \exp \left( -\frac{(g_x - x)^2 + (g_y - y)^2 }{\sigma_{\mgps}^2} \right)
\end{equation}
where $g_x$ and $g_y$ is the GPS observation converted from Universal
Transverse Mercator (UTM) coordinate to map coordinate.

\paragraph{Vehicle Motion Model:}
Our motion model encodes the
fact that the inferred vehicle velocity should agree with the
vehicle dynamics, given previous time's belief. In particular, wheel odometry and IMU are used as input to an extended Kalman filter to generate an estimate of the velocity of the vehicle. 
We then define the motion model  to be
\begin{equation}
  \label{eq:motion-model}
  \mathrm{Bel}_{t|t-1} (\bx | \mathcal{X}_t) = \sum_{\bx_{t-1} \in
  \mathcal{R}_{t-1}} P(\bx | \mathcal{X}_t, \bx_{t-1}) \mathrm{Bel}_{t-1}(\bx_{t-1})
\end{equation}
where
\begin{equation} \label{eq:dynamics-energy}
P(\bx | \mathcal{X}_t, \bx_{t-1})  \propto \rho \left(\bx \ominus (\bx_{t-1} \oplus \mathcal{X}_t)\right),
\end{equation}
with $\rho = \exp\left( -\bz^T \Sigma^{-1} \bz  \right)$ is a Gaussian error function.
$\Sigma$ is
the covariance matrix and $\mathcal{R}_{t-1}$ is our three-dimensional search range centered at previous step's $\bx_{t-1}^\ast$. 
$\oplus$, $\ominus$ are the 2D pose composition operator and inverse pose
composition operator respectively, which,
following~\citet{kummerle2009measuring} are defined as
\[
\mathbf{a} \oplus \mathbf{b} = 
\left[\begin{array}{c}
x_a + x_b \cdot \cos\theta_a - y_b \cdot \sin\theta_a \\
y_a + x_b \cdot \sin\theta_a + y_b \cdot \cos\theta_a\\
\theta_a + \theta_b 
\end{array}\right], 
\mathbf{a} \ominus \mathbf{b} = 
\left[\begin{array}{c}
(x_a - x_b) \cdot \cos\theta_b + (x_b - y_b) \cdot \sin\theta_b \\
-(x_a - x_b) \cdot \sin\theta_b + (x_b - y_b) \cdot \cos\theta_b \\
\theta_a - \theta_b 
\end{array}\right].
\]

\paragraph{Network Architectures:}
The $f$ and $g$ functions  are computed via multi-layer fully
convolutional neural networks. We experiment with a 6-layer network based on
the patch matching architecture used by \citet{luo2016efficient} and with
LinkNet by \citet{linknet}.
We use instance normalization~\citep{instancenorm} after each
convolutional layer \slwang{instead of batch norm, due to its capability of 
reducing instance-specific mean and covariance shift. Our embedding output has
the same resolution as the input image, with a (potentially) multi-dimensional embedding per pixel. The dimension is chosen based on the trade-off between performance and runtime. } %
We refer the reader to \fref{fig:embedding-example}
 for an illustration of a single-channel embedding. All our experiments use
 single-channel embeddings for both online \lidar{}, as well as the maps,
 unless otherwise stated. 

\subsection{Online Localization}
Estimating the pose of the \andreibc{vehicle} at each time step $t$ requires solving the
maximum a posteriori problem:
\begin{equation}
\bx_t^\ast = \arg\max_\bx \mathrm{Bel}_t(\bx) = \arg\max_\bx \eta \cdot
P_{\mlidar}(\cI_t | \bx; \bw) P_{\mgps}(\mathbf{g}_t | \bx) \mathrm{Bel}_{t|t-1}
(\bx).
\end{equation}
This is a complex inference over the continuous variable $\bx$, which is non-convex and requires intractable integration. 
These types of problems are typically solved with sampling approaches such as
particle filters,  which can easily fall into local minima. 
Moreover, most particle solvers have non-deterministic run times, which is
problematic for safety-critical real-time applications like self-driving cars.

Instead, we follow the histogram filter approach to compute $\bx_{t}^\ast$ through a search-based method
which is much more efficient, given the characteristics of the problem.
To this end, we discretize the 3D search space over
$\bx = \{ x, y, \theta \}$ as a grid, and compute the term $\mathrm{Bel}_t(\bx)$ for
every cell of our search space.
We center the search space at the so-called dead reckoning pose $\dot{\bx}_{t|t-1} = \arg \max_\bx \mathrm{Bel}_{t|t-1}(\bx)$, which
represents the pose of the vehicle at time $t$  estimated  
using  IMU and wheel encoders.
\andreibc{Inference happens in the vehicle coordinate frame, with $x$ being the
longitudinal offset along the car's trajectory, $y$, the latitudinal offset,
and $\theta$ the heading offset.}
\slwang{The search range is selected in order to find a compromise between the
computational cost and the capability of handling large drift. }

In order to do inference in real-time, we need to compute each term efficiently.
The GPS term $P_{\mgps}(\mathbf{g}_t | \bx)$ is a simple Gaussian kernel.
The motion $\mathrm{Bel}_{t|t-1}(\bx)$ computation is quadratic w.\ r.\ t.\ the
number of discretized states.  Given the fact that it is a small neighborhood
around the dead reckoning pose, the computation is very fast in practice.
The most computationally demanding component of our model is
the fact that we need to enumerate all possible locations in our search
range to compute the \lidar{} matching term $P_{\mlidar}(\cI_t | \bx; \bw)$. 
However, we observe that computing the inner product scores between
two 2D deep embeddings across all translational positions in our $(x, y)$
search range is equivalent to convolving the map embedding with the online
embedding as a kernel. This makes the search over $x$ and $y$ much faster to
compute. As a result, the entire optimization of $P_{\mlidar}(\cI_t | \bx; \bw)$ can be
performed using $n_\theta$ convolutions, where $n_\theta$
is the number of discretization cells in the rotation ($\theta$) dimension.

State-of-the-art GEMM/Winograd based \andreibc{(spatial)} GPU convolution
implementations are often optimized for small convolutional kernels. Using
these for the GPU-based convolutional matching implementation is
still \andreibc{too slow for our real-time operation goal}.
This is due to the fact that our ``convolution kernel'' \andreibc{(i.e., the
online embedding)} is very large in
practice (in our experiments the size of our online \lidar{} embedding is
$600\times 480$\andreibc{, the same size as the online \lidar{} image}).
In order to speed this up, we perform this operation
in the Fourier domain, as opposed to the spatial one, according to convolution
theorem: $f \ast g = \cF^{-1}(\cF(f) \odot \cF(g))$, where ``$\odot$'' denotes
an element-wise product.
This reduces the \andreibc{theoretical} complexity from $O(N^2)$ to $O(N\log N)$,
which translates to massive improvements in terms of run time, as we will
show in Section~\ref{sec:result}.

Therefore, we only need to run the embedding networks once, rotate the
computed online \lidar{} embedding $n_\theta$ times, and convolve each
rotation with the map embedding to get the probability for all the pose
hypotheses in the form of a score map $S$. Our 
solution is therefore globally optimal over our discretized
search space  including both rotation and translation. In practice,
the rotation of our online \lidar{} embedding is implemented using a spatial
transformer module~\cite{jaderberg2015spatial}, and generating all rotations
takes 5ms in total (we use $n_\theta = 5$ in all our experiments).

In order to handle robustness to observation noise and bring smooth
localization results to avoid sudden jumps, we exploit a soft version of the
argmax \cite{levinson2010robust}, which is a trade-off between center of mass and argmax:
\begin{equation}
\label{eq:soft-argmax}
\bx_t^\ast= \frac{\sum_\bx \mathrm{Bel}_t(\bx)^\alpha \cdot \bx}{\sum_\bx \mathrm{Bel}_t(\bx)^\alpha}
\end{equation}
where $\alpha$ is a temperature hyper-parameter larger than 1. This gives us an
estimation that takes the uncertainty of the prediction into account at time
$t$.

\subsection{Learning}

The localization system is end-to-end differentiable, enabling us to learn
all parameters jointly using back-propagation. We find that a simple cross-entropy
loss is sufficient to train the system, without requiring any additional, potentially
expensive terms, such as a reconstruction loss.
We define the cross-entropy loss between the ground-truth position and the
inferred score map as 
$
   \cL = -\sum_i \bp_{i, \text{gt}} \log \bp_i,
$
where the  labels $\bp_{i, \text{gt}}$ are represented as
one-hot encodings of the ground truth position, i.e., a tensor with
the same shape as the score map $S$, with a 1 at the correct pose.

\begin{figure*}[tb]
\centering
\setlength{\tabcolsep}{1pt}
\begin{tabular}{ccc}
\includegraphics[width=0.33\linewidth, trim={1cm 0 0.5cm
0}]{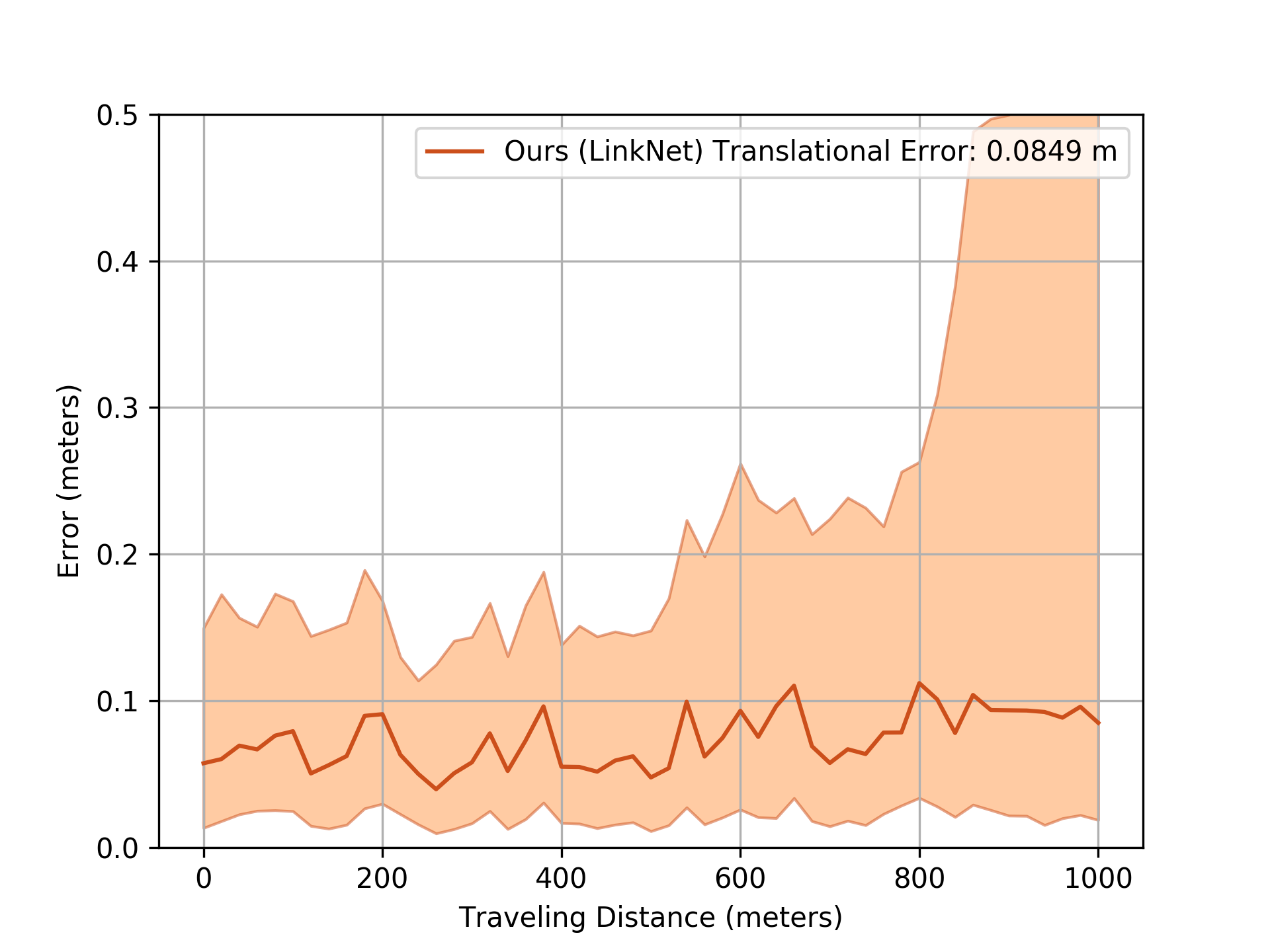}
& \includegraphics[width=0.33\linewidth, trim={1cm 0 0.5cm
0}]{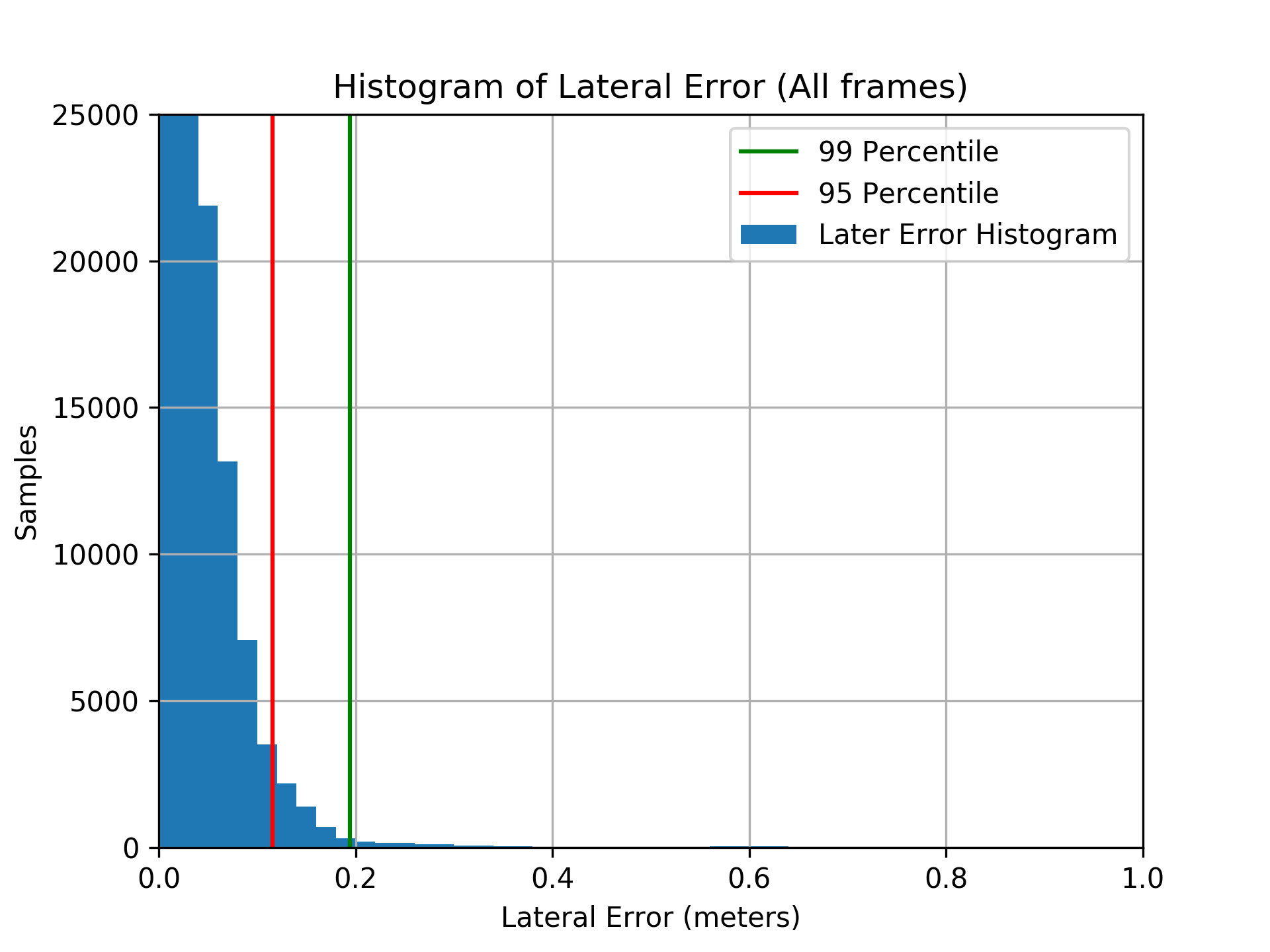}
& \includegraphics[width=0.33\linewidth, trim={1cm 0 0.5cm
0}]{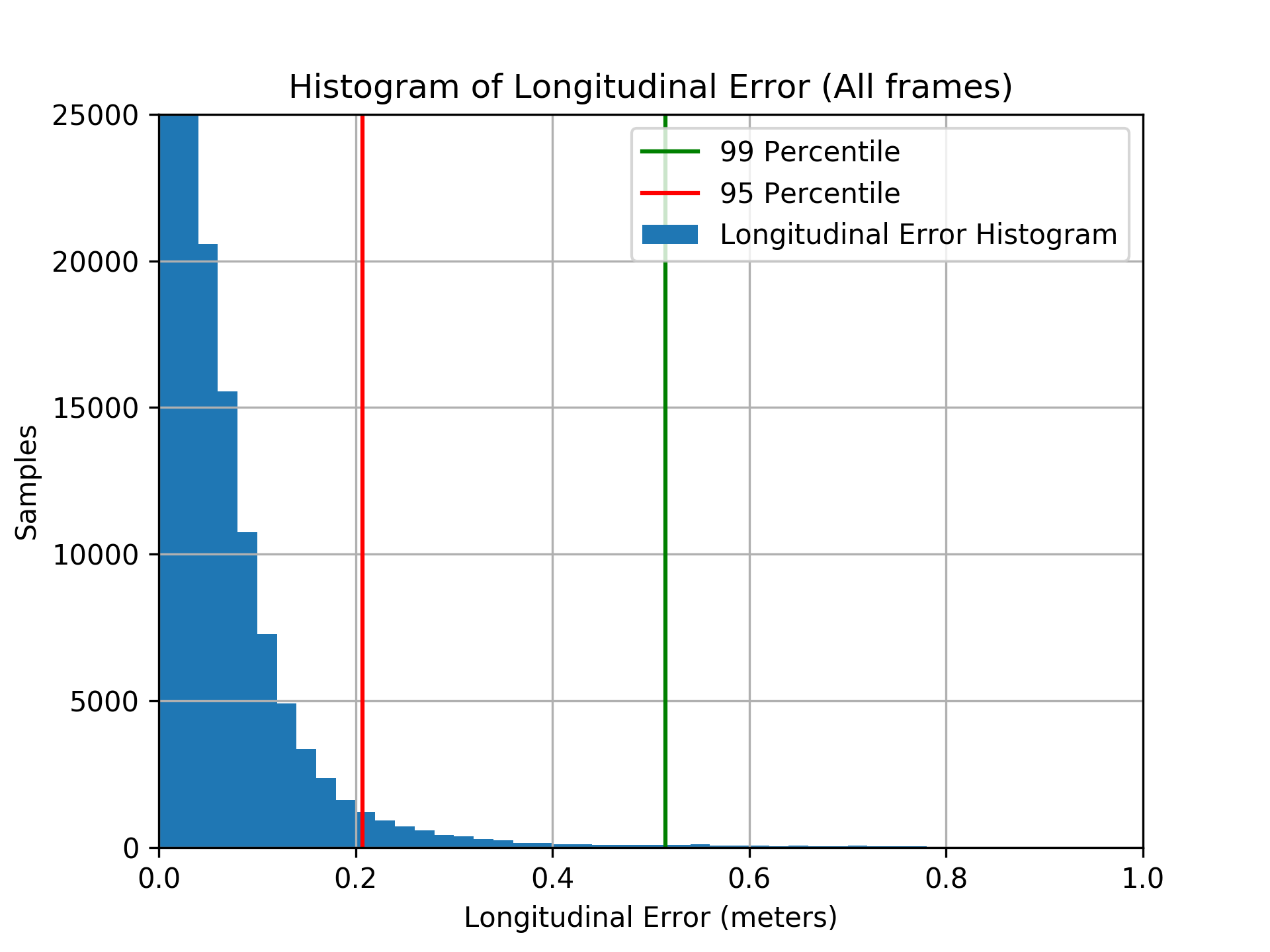} \\
Error vs Traveling Dist & Lateral Histogram & Longitudinal Histogram
\\\end{tabular}
\caption{Quantitative Analysis. From left to right: localization error vs traveling distance; lateral error histogram per each timestamp, longitudinal histogram per each step. }
\label{fig:qual-comparison}
\end{figure*}

\begin{figure*}[t]
\centering
\setlength{\tabcolsep}{1pt}
\begin{tabular}{ccc}
\includegraphics[width=0.33\linewidth, trim={0 0 0 0}]{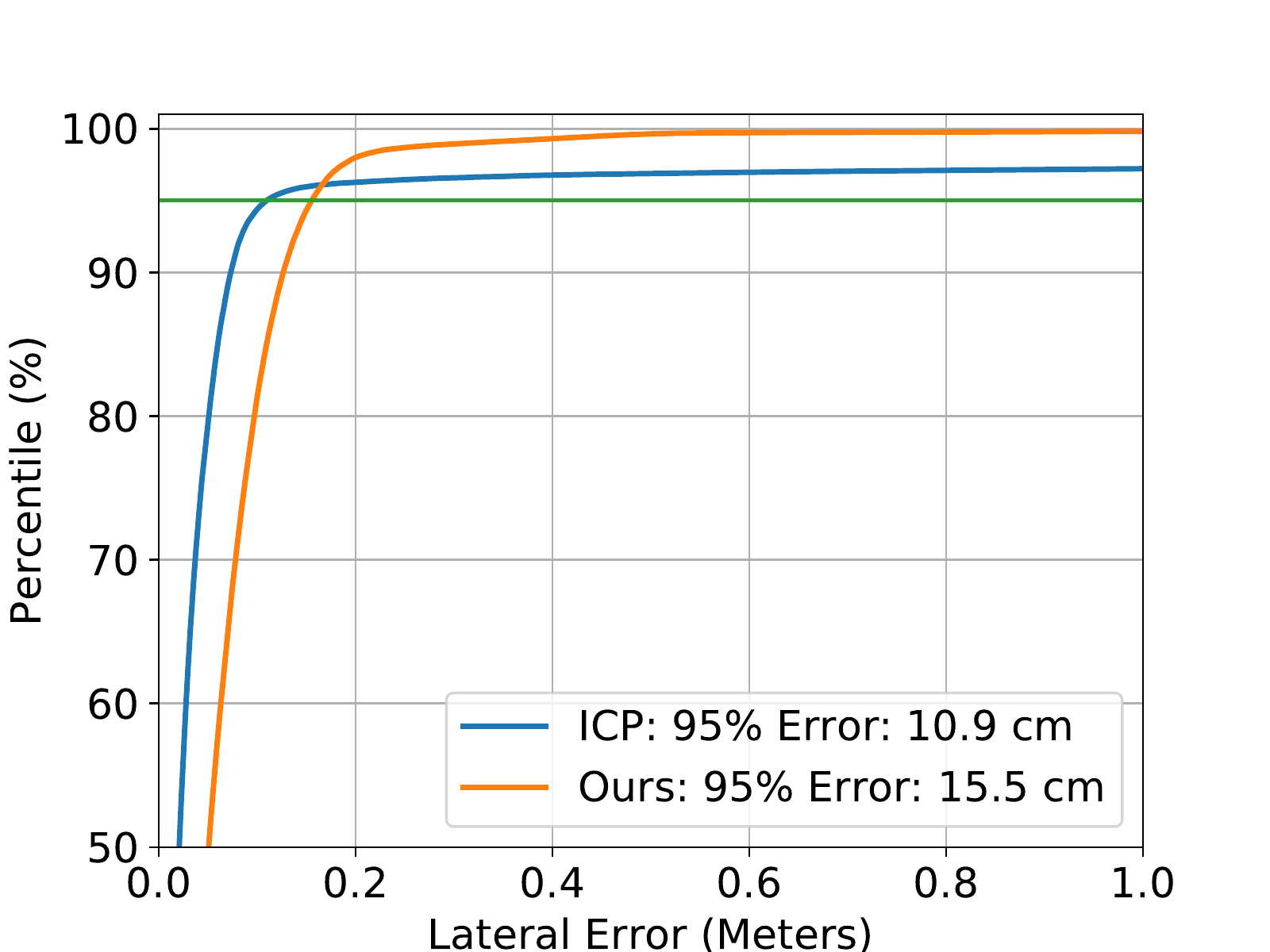}
& \includegraphics[width=0.33\linewidth, trim={0 0 0 0}]{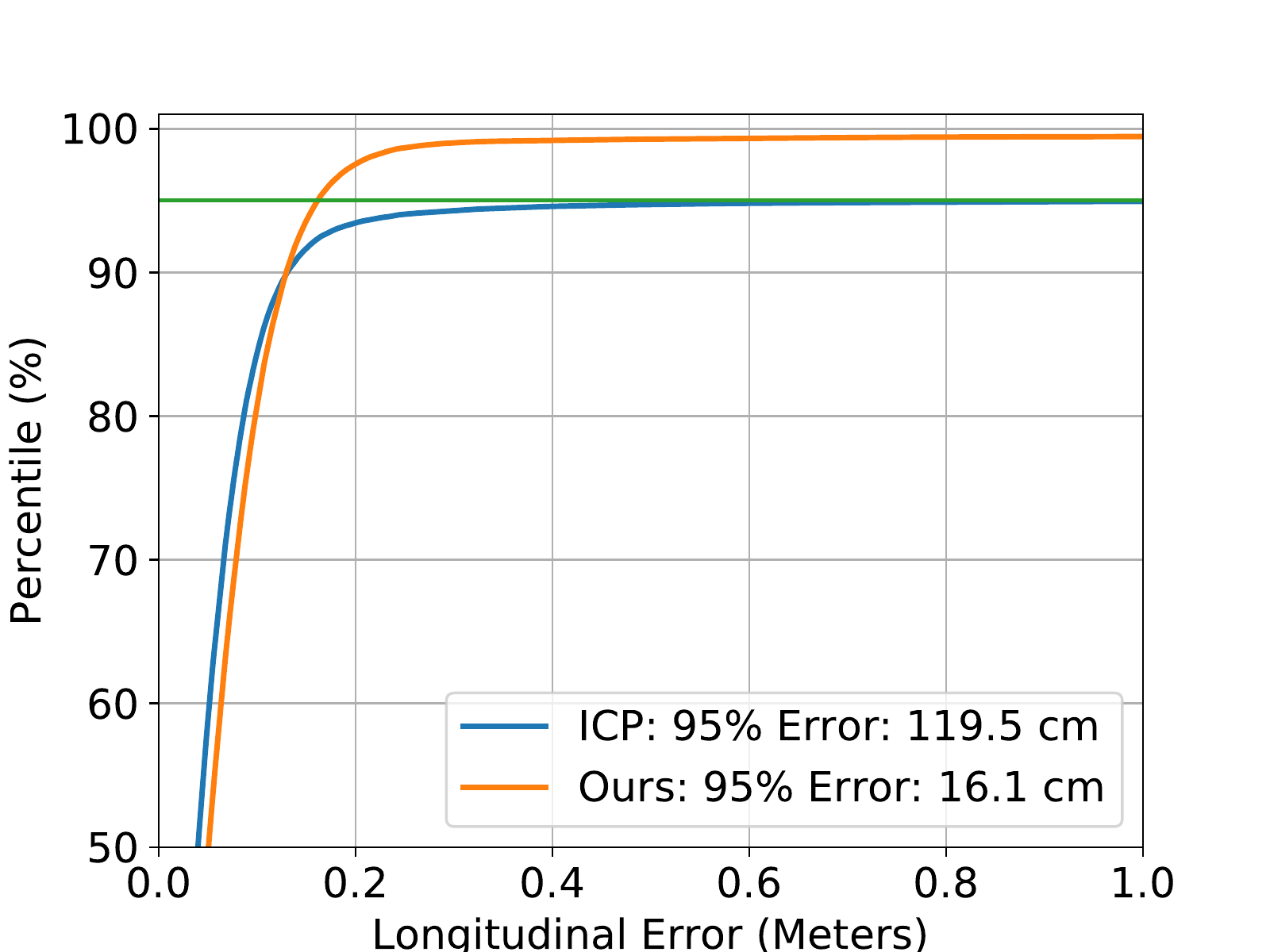}
& \includegraphics[width=0.33\linewidth, trim={0 0 0 0}]{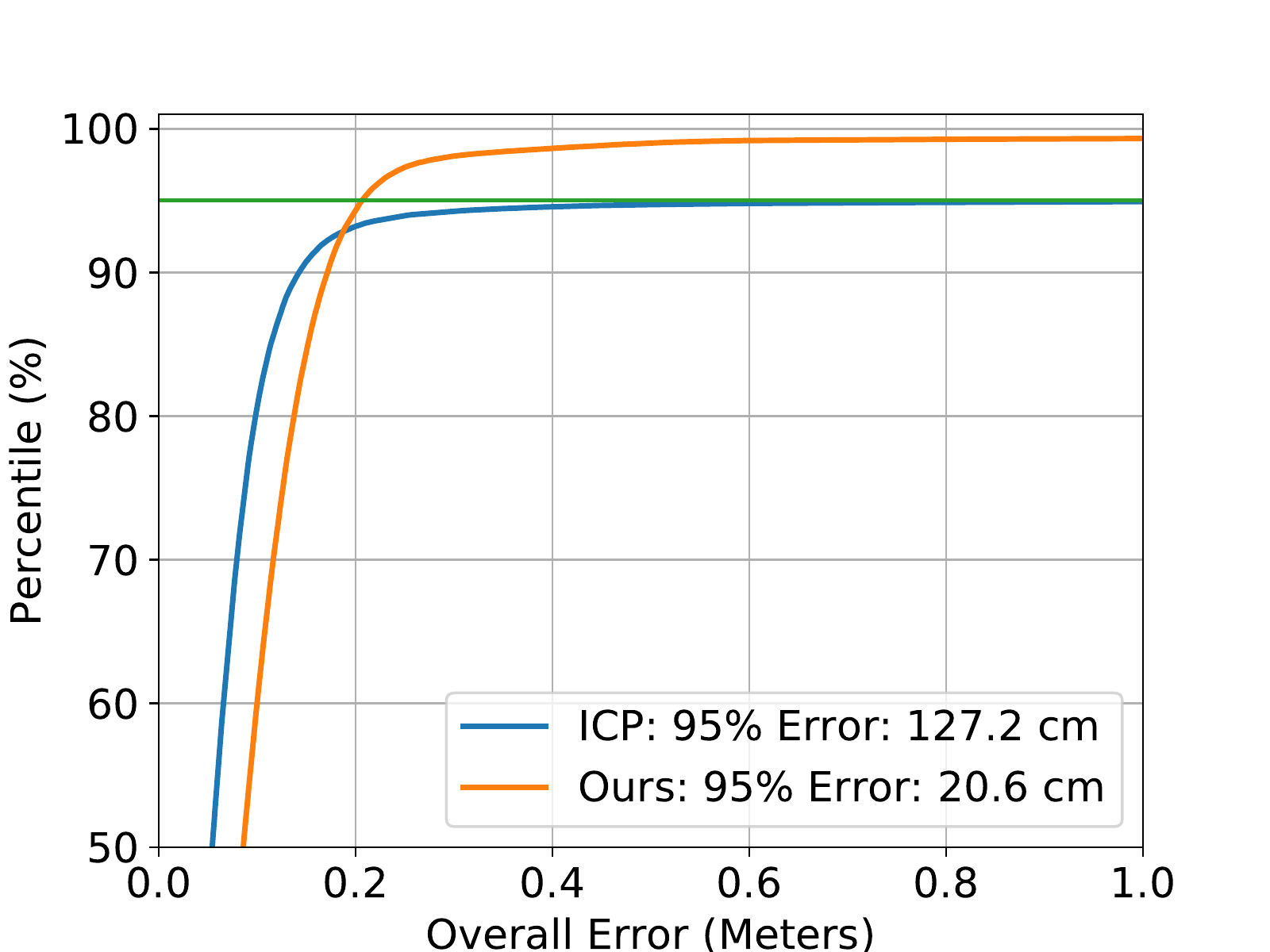} \\

Lateral & Longitudinal & Total Translational
\\\end{tabular}
\caption{Cumulative error curve. From left to right: lateral, longitudinal, total translational error. }
\label{fig:qual-cum}
\vspace{-5mm}
\end{figure*}

\section{Experimental Results}
\label{sec:result}

\begin{figure*}[t]
\centering
\setlength{\tabcolsep}{1pt}
\begin{tabular}{ccc}
\multicolumn{3}{c}{\includegraphics[width=0.98\linewidth, trim={0 0 0 0}]{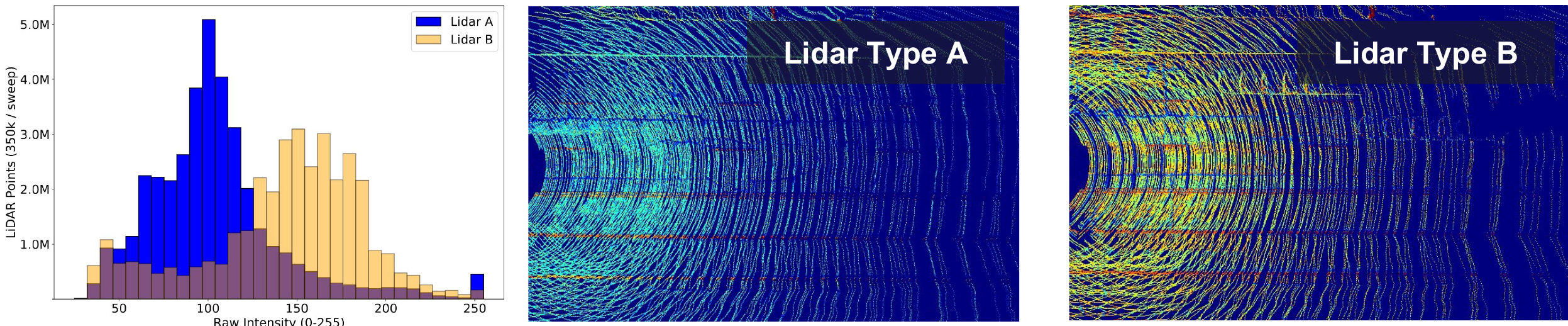}} \\   
\\\end{tabular}
\vspace{-0.25cm}
  \caption{A comparison between the two \lidar{} sensors: left: the different intensity profiles of their sweeps over the same location; right: the color-mapped intensity image.}
  \label{fig:sensor-transfer}
\end{figure*}

\paragraph{Dataset:}
We collected a new dataset comprising over 4,000km of
driving through a variety of urban and highway environments in 
multiple cities/states in North America, collected with two types of \lidar{}
sensors. According to the scenarios we split our dataset into
\textit{Highway-LidarA} and \textit{Misc-LidarB}, where \textit{Highway-LidarA}
contains over 400 sequences for a total of over 3,000km of driving for
training and validation. \andreibc{We select a representative and challenging subset of 
282km  of driving for testing, ensuring that there is no geographic overlap
between the splits.}
All these sequences are collected by a \lidar{} type A.
\textit{Misc-LidarB} contains 79 sequences with 200km of driving 
over a mix of highway and city  collected by
a different \lidar{} type B in a different state. \lidar{}s A and B differ
substantially in their intensity output profiles, as shown in
~\fref{fig:sensor-transfer}.

\renewcommand{\arraystretch}{1.1}

\paragraph{Experimental Setup:}
We randomly extracted 230k training samples from the training sequences.
For each training sample, we aggregate the five most recent online \lidar{} sweeps to generate the BEV intensity image using vehicle dynamics, corresponding to 0.5 seconds of \lidar{} scan. In such a short time drift is negligible. 
Our ground-truth poses are acquired
through an expensive high precision offline matching procedure with up to several centimeter uncertainty.
We rasterize the aggregated \lidar{} points to create a \lidar{} intensity
image. Both the online intensity image and the intensity map are discretized at
a spatial resolution of 5cm covering a $30m\times24m$ region. During matching, we use the same spatial
resolution, plus a rotational resolution of 0.5\degree, with a total search 
range of $1m\times{}1m\times{}2.5$\degree around the dead reckoning pose.
We report the median error as well as the
failure rate. 
The median error  reflects how accurate the localization is
in the majority of cases while the failure rates reflect the worst case
performance
In particular, we define  ``failure'' if there is at least a frame with
localization error over \andreibc{1m}. In addition to these per-sequence
metrics, we also plot the per-frame cumulative localization error curve in
\fref{fig:qual-cum}.

\renewcommand{\arraystretch}{1.2}

\begin{table}[t]
  \centering
  \caption{Localization Performance on \textbf{Highway-LidarA} Dataset (Per
  Sequence) \label{tab:main-quantitative}. Please note that the numbers in this
version of the paper are more up to date than those in the CoRL proceedings
after a small bug was fixed.}
   \begin{tabular}{ccc|ccc|ccc}
\toprule
     & &  & \multicolumn{3}{|c|}{\textbf{Median Error (cm)}}
     & \multicolumn{3}{|c}{\textbf{Failure Rate (\%)}} \\
     \textbf{Method} &$\mathrm{Motion}$  & $\mathrm{Prob}$ & Lat & Lon
                     & Total
                     & $\le 100$m & $\le 500$m &$\le$ End
     \\\midrule
Dynamics & Yes & No & 439.21 & 863.68 & 1216.01 & 0.46 & 98.14 & 100.00  \\
Raw \lidar{} & Yes & No & 1245.13 & 590.43 & 1514.42 & 1.84 & 81.02 & 92.49  \\
ICP & Yes &  No & \textbf{1.52} & 5.04 & \textbf{5.44} & 3.50 & 5.03 & 7.14  \\
\midrule

Ours (LinkNet) & No & No & 3.87   & 4.99 & 7.76  & 0.35 & 0.35 & 0.72  \\
Ours (LinkNet) & Yes & No & 3.81  & 4.53 & 7.18 & 1.06 & 1.06 & 1.44 \\
Ours (LinkNet) & Yes & Yes & 3.00 & \textbf{4.33} & 6.47
               & \textbf{0.00} & \textbf{0.00} & \textbf{0.00}  \\
\bottomrule
     \end{tabular}
\end{table}

\begin{table}[t]
  \centering
  \caption{Localization Performance on \textbf{Misc-LidarB} trained on \textbf{Highway-LidarA} (Per Sequence) \label{tab:main-transfer}}
   \begin{tabular}{ccc|ccc|ccc}
\hline
     & & & \multicolumn{3}{|c|}{\textbf{Median Error (cm)}}
     & \multicolumn{3}{|c}{\textbf{Failure Rate (\%)}} \\
\textbf{Method} & $\mathrm{Motion}$ & $\mathrm{Prob}$ & Lat & Lon & Total
                     & $\le 100$m & $\le 500$m &$\le$ End
     \\ \hline

     Dynamics Only & Yes & No & 195.73 & 322.31 & 468.53 & 6.13 & 68.66 & 84.26  \\
     ICP & Yes & No & \textbf{2.57} & 15.29 & 16.42 & 0.46 & 28.43 & 37.53  \\ \hline
     Ours (Transfer) & Yes & No & 6.95 & \textbf{6.38} & \textbf{11.73} & \textbf{0.00}
                  & \textbf{0.71} & \textbf{1.95}  \\
  \hline
     \end{tabular}
\end{table}

\paragraph{Implementation Details:} 

\slwang{We manually chose the following hyper-parameters through validation,
namely the motion model variance $\Sigma=\mathrm{diag}([3.0, 3.0, 3.0])$, GPS's
observation variance $\sigma_{gps}=10.0$, temperature constant $\alpha=2.0$. We
also conduct two ablation studies. Our first ablation verifies whether the
motion prior defined in Eq.~\eqref{eq:motion-model} is helpful. 
We evaluate algorithm with and without this term, denoted as $\mathrm{Motion}$
in~\tref{tab:main-quantitative}. Our second ablation evaluates whether
a probabilistic MLE proposed in Eq.~\eqref{eq:soft-argmax} helps improve
performance, denoted as $\mathrm{Prob}$. The none-probabilistic is achieved
through changing the soft-argmax in Eq.~\eqref{eq:soft-argmax} to a hard argmax. We implement our full inference algorithm in PyTorch 0.4.  The networks are trained using Adam over four NVIDIA 1080Ti GPUs with initial learning rate at 0.001.  }

\paragraph{Comparison to Other Methods:}
We compare our algorithm against several baselines. The raw matching consists of performing the
matching-based localization in a similar manner to our  method but only use the
raw intensity BEV online and map images, instead of the learned
embeddings. The ICP baseline conducts point-to-plane ICP between the raw 3D
\lidar{} points against the 3D pre-scanned localization prior at 10Hz, initialized in
a similar manner as us using the previous estimated location plus the vehicle
dynamics. \andreibc{This ensures good quality of initialization, as
required by algorithms from ICP family.}

\paragraph{Localization Performance:}
As shown in  \tref{tab:main-quantitative}, our
approach achieves the best performance among all the competing
algorithms in terms of failure rate.  Both probabilistic inference and motion
prior further improves the robustness of our method.
Our ICP baseline is competitive in terms of median
error, especially along lateral direction, but the failure rate is
significantly higher.  It is also more computationally demanding and requires 3D
maps. Both dynamics-only and raw intensity matching result in large drift. Moreover, we have observed that deeper architectures
and the probabilistic inference are generally helpful. 
\fref{fig:qual-comparison} shows the localization error as
a function of the travel distance aggregated across all sequences from the
\textit{Highway-LidarA} test set. The solid line denotes the median and the shaded region denotes the $95\%$ area, together with the distribution of lateral and
longitudinal errors per frame. 
\fref{fig:qual-cum} compares our approach to ICP in terms of cumulative errors
with $95\%$-percentile error reported. From this we can see our method
significantly outperforms ICP in terms of the worst-case behavior. %

\paragraph{Domain Shift:}
In order to show that our  approach generalize well across
  \lidar{} sensors, we conduct a second experiment, where we train our network
  on \textit{Highway-LidarA}, which is purely highway, collected using \lidar{} A,
and test on the test set of \textit{Misc-LidarB} which is highway + city,
collected by a different \lidar{} (type B) in a different state.
In order to better highlight the difference, in \fref{fig:sensor-transfer} we
show two \lidar{}s intensity value distributions and their raw intensity images, 
collected at the same location.
\tref{tab:main-transfer} showcases \andreibc{the results of this experiment}.
From the table we can see our neural network is able to generalize
both across \lidar{} models and across environment types.

\paragraph{Runtime Analysis:}
\slwang{We conduct \andreibc{a} runtime analysis over both embedding networks and matching. 
Our LinkNet based embedding networks take less than 10ms each for a forward pass over both online
and map images. We also compare the \cudnn{} implementation of FFT-conv and standard spatial convolution. 
\andreibc{FFT reduces the run time of the matching \andreibc{by an order of
magnitude} bringing it down from $27ms$ to $1.4ms$}
for a \andreibc{single}-channel embedding.
This enables us to run the localization algorithm at 15 Hz, thereby 
achieving our real-time operation goal.}

\section{Conclusion}
\label{sec:conclusion}

We proposed a real-time, calibration-agnostic, effective localization method
for self-driving cars.  Our method projects the online \lidar{} sweeps and
  intensity map into a joint embedding space. Localization is conducted through
  efficient convolutional matching between the embeddings. This approaches
  allows our full system to operate in real-time at 15Hz while achieving
  centimeter-level accuracy without intensity calibration. The method also
  generalizes well to different \lidar{} types without the need to re-train. 
  The experiments
  illustrate the performance of the proposed approach over two comprehensive
  \andreibc{test sets covering over 500km of driving in diverse conditions}.

\acknowledgments{We would like to thank Min Bai, Julieta Martinez, Joyce Yang,
  and Shrinidhi Kowshika Lakshmikanth for their help with proofreading,
  experiments, and dataset generation.
We would also like to thank our anonymous reviewers for the detailed feedback
they provided on our work, which helped us improve our system and paper in
multiple ways.}

\bibliography{refs}  %

\clearpage{}

\appendix

\section{Supplementary Material}
\label{sec:supmat}

\subsection{Additional Ablation Study Results}

\paragraph{Embedding Dimensions:}
\slwang{
We also investigate the impact that the number of embedding channels has on matching performance and the runtime of the system. 
\tref{tab:ablation-channels} shows the performance. As shown in this table,
increasing the number of channels in the embeddings does not improve performance by a significant
amount, whereas reducing the number of channels could reduce the runtime of
matching by a large margin, which favors relative low-dimensionality in
practice.} \andreibc{Therefore, using single-channel embeddings (just like the input
intensity images) is adequate.}

\paragraph{Network Architectures}
We experiment with two configurations of
embedding networks. The first one,
denoted as FCN, uses the shallow network described
in Section~\ref{sec:robust-localization} for both online and map branches. The second architecture uses a LinkNet~\cite{linknet} architecture for both the online and the
map branches. The results are reported in \tref{tab:architecture}. From the
table, we can see LinkNet achieves better performance than FCN and the motion
model consistently helps when using either architecture.

\paragraph{Reduced Training Dataset Size:}
Given that the matching task is conceptually straightforward, requiring far
less high-level reasoning capabilities compared to problems such as semantic
segmentation, we perform a series of experiments where we train our matching
network using smaller samples of our training dataset, and investigate the
localization performance of our system in these cases. These results are
presented in~\tref{tab:ablation-dataset-lidar-a}.

\begin{table}[bpht!]
  \centering
  \caption{Localization performance under varying numbers of embedding
    channels, as measured on an NVIDIA GeForce GTX 1080 Ti GPU running CUDA
    9.2.88 and cuDNN 7.104 on driver version 396.26. The matching accuracy
      represents the percentage of \andreibc{predictions within one pixel of
    the ground truth. Results averaged over 500 forward passes.} \label{tab:ablation-channels}}
   \begin{tabular}{lrrrrrrrr}
     \toprule
       \textbf{Method}
       & \textbf{\shortstack{Matching Accuracy}}
     & \phantom{x}
     & \multicolumn{3}{c}{\textbf{Inference Time (ms)}}
     \\
     \cmidrule{4-6} & & & Backbones & Matching (Slow)
                                 &  Matching (FFT) \\
     \midrule
      Raw matching & 13.95\% & & n/A	 	   & 26.66ms 	 & 1.43ms  	 \\
      1 channel 	 & 71.97\% & & 19.34ms 	 & 26.66ms 	 & 1.43ms  	 \\
      2 channels 	 & 72.08\% & & 17.16ms 	 & 55.03ms 	 & 6.60ms  	 \\
      4 channels 	 & 71.67\% & & 16.70ms 	 & 110.46ms 	 & 11.18ms  	 \\
      8 channels 	 & 71.63\% & & 18.08ms 	 & 168.96ms 	 & 21.64ms  	 \\
      12 channels  & 72.50\% & & 18.73ms 	 & 330.75ms 	 & 32.93ms  	 \\
     \bottomrule
     \end{tabular}
\end{table}

\begin{table}[bpht!]
  \centering
  \caption{Localization performance using different backbone architectures on
  our \textbf{Highway-LidarA} dataset.\label{tab:architecture}}
   \begin{tabular}{ccc|ccc|ccc}
\toprule
& & & \multicolumn{3}{|c|}{\textbf{Median Error (cm)}}
     & \multicolumn{3}{|c}{\textbf{Failure Rate (\%)}} \\
\textbf{Method} &$\mathrm{Motion}$ & $\mathrm{Prob} $  & Lat & Lon
                     & Total
                     & $\le 100$m & $\le 500$m &$\le$ End
     \\\midrule

     Ours (FCN) & No & No   & 4.41 & \textbf{4.86} & 8.01 & 0.35 & 0.35 & 0.71  \\
  Ours (FCN) & Yes & Yes & 5.50 & 6.00 & 9.52 & 1.06 & 1.42 & 2.52  \\

  Ours (LinkNet) & No & No & 3.87   & 4.99 & 7.76  & 0.35 & 0.35 & 0.72  \\
  Ours (LinkNet) & Yes & Yes & \textbf{3.19} & \textbf{4.86} & \textbf{7.09}
                 & \textbf{0.00} & \textbf{0.00} & \textbf{0.35}  \\
\bottomrule
     \end{tabular}
\end{table}

\begin{table}[bpht!]
  \centering
  \caption{Localization performance using a matching network trained on less
  data on our Highway-LidarA dataset.}
  \label{tab:ablation-dataset-lidar-a}
   \begin{tabular}{ccc|ccc|ccc}
\toprule
& & & \multicolumn{3}{|c|}{\textbf{Median Error (cm)}}
     & \multicolumn{3}{|c}{\textbf{Failure Rate (\%)}} \\
\textbf{Method} & $\mathrm{Motion}$ & $\mathrm{Prob}$ & Lat & Lon
                     & Total
                     & $\le 100$m & $\le 500$m &$\le$ End \\
\midrule
LinkNet, 100\% of data & Yes & Yes & 3.19 & \textbf{4.86} & \textbf{7.09}
                 & \textbf{0.00} & \textbf{0.00} & \textbf{0.35}  \\
LinkNet, 25\% of data & Yes & Yes & \textbf{2.92} &	5.33 &	7.24
                      &	\textbf{0.00} &	\textbf{0.00} &	1.08 \\ 
LinkNet, 5\% of data & Yes & Yes & 3.95 &	6.76 &	9.25 &	1.06 &	1.06 &	2.52
\\
LinkNet, 1\% of data & Yes & Yes & 4.66 &	8.70 & 11.40 &	0.71 &	2.14 &	3.60 
\\
\bottomrule
  \end{tabular}
\end{table}

\subsection{Additional Qualitative Results}

\subsection{Qualitative Results}

\begin{figure*}[bpht!]
  \centering
  \begin{subfigure}[t]{0.98\textwidth}
    \begin{center}
      \includegraphics[width=\textwidth]{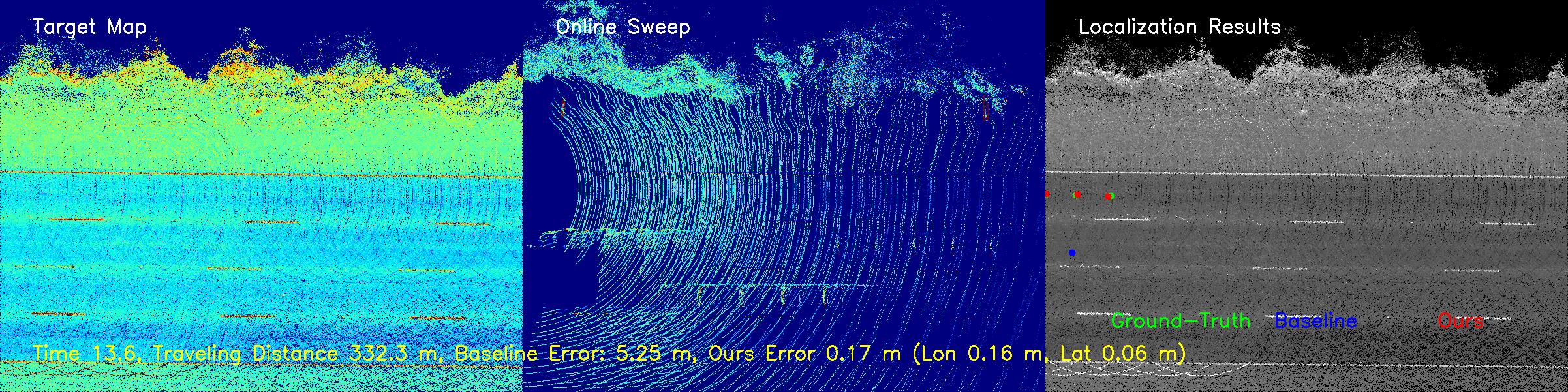}
      \caption{Repetitive geometric structures on a highway (challenging to
      localize longitudinally with a pure geometric method).}
    \end{center}
  \end{subfigure}
  \begin{subfigure}[t]{0.98\textwidth}
    \centering
    \includegraphics[width=\textwidth]{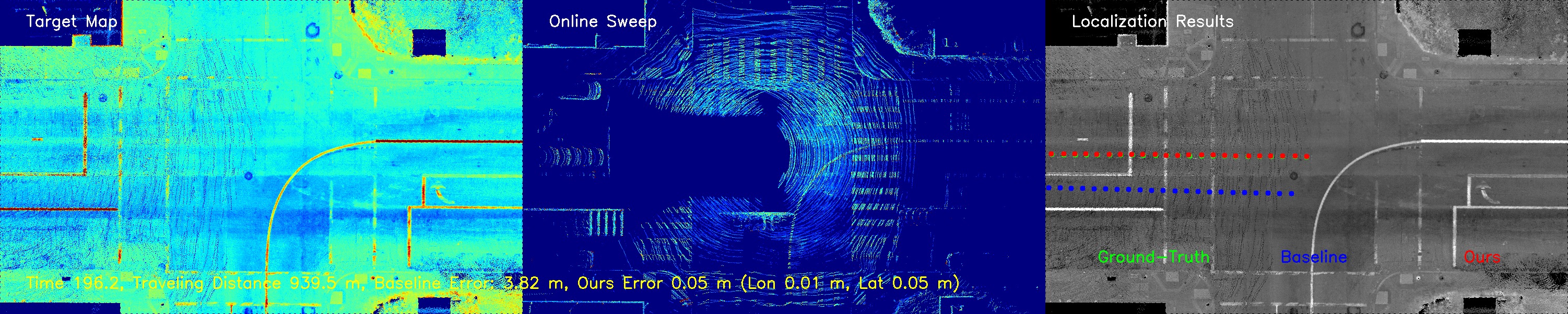}
    \caption{Changes in road markings (note the different pedestrian crossing
    markings in the map vs.\ the perceived online \lidar{}).}
  \end{subfigure}
  \begin{subfigure}[t]{0.98\textwidth}
    \centering
    \includegraphics[width=\textwidth]{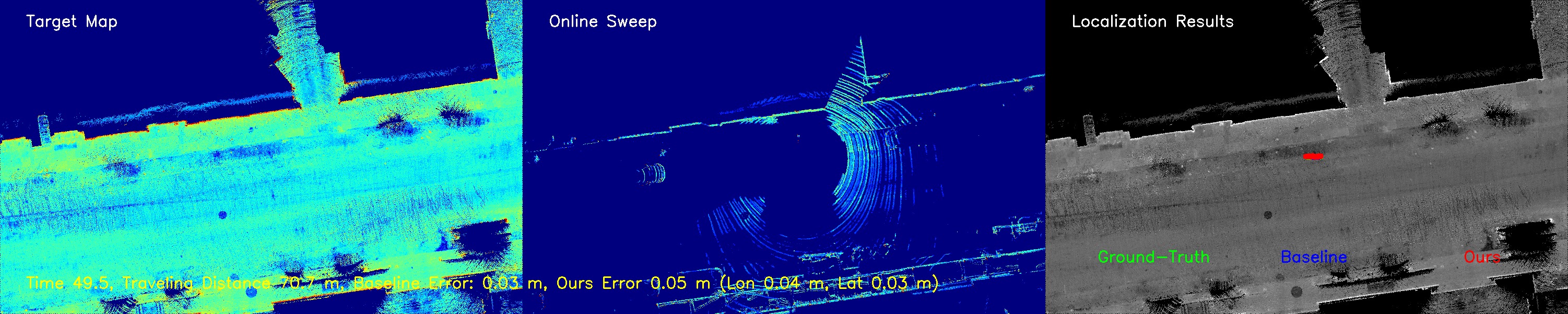}
    \caption{Reverse parallel parking.}
  \end{subfigure}
  \begin{subfigure}[t]{0.98\textwidth}
    \centering
    \includegraphics[width=\textwidth]{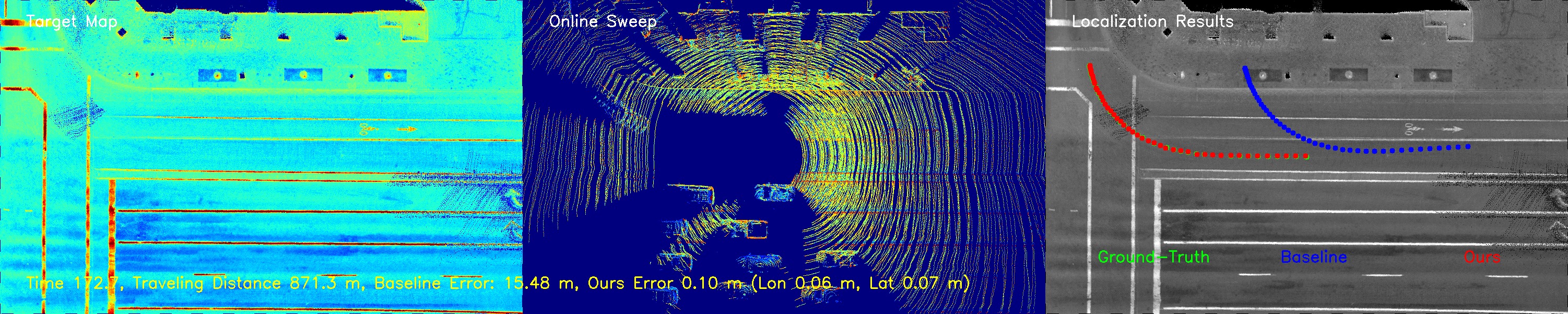}
    \caption{A sharp turn into an intersection.}
  \end{subfigure}

 \caption{Qualitative examples of several interesting scenarios in which our
    system is able to localize successfully. Here, just like in our video, the
  method labeled as ``baseline'' is the dynamics-only baseline.}
    \label{fig:comparison}
\end{figure*}

\fref{fig:comparison} qualitatively compares the
localization accuracy of our method with those discussed in the previous
section. For further qualitative results, please refer to the video associated
with this paper.

\end{document}